\documentclass[11pt]{article}

\usepackage[preprint]{acl}

\usepackage{times}
\usepackage{latexsym}

\usepackage[T1]{fontenc}

\usepackage[utf8]{inputenc}

\usepackage{microtype}

\usepackage{inconsolata}
\usepackage{booktabs}
\usepackage{xspace}          
\usepackage{amsmath}         
\usepackage{amssymb}         
\usepackage{algorithm}       
\usepackage{algpseudocode}   
\usepackage{tcolorbox}
\tcbuselibrary{listings, breakable, skins}
\usepackage{listings}
\usepackage[table]{xcolor}  

\usepackage{xcolor}
\newcommand{\tablestyle}[2]{\setlength{\tabcolsep}{#1}\renewcommand{\arraystretch}{#2}\centering\footnotesize}
\usepackage{pifont}

\lstdefinestyle{promptstyle}{
  basicstyle=\ttfamily\footnotesize,
  breaklines=true,
  breakatwhitespace=true,
  columns=fullflexible,
  showstringspaces=false,
  aboveskip=0pt,
  belowskip=0pt
  breakindent=0pt,
breakautoindent=false,
}

\newtcolorbox{promptbox}[1]{
  enhanced,
  breakable,
  colback=white,
  colframe=black!20,
  boxrule=0.5pt,
  arc=2pt,
  left=4pt,right=4pt,top=4pt,bottom=4pt,
  title=\textbf{#1},
  fonttitle=\small,
}
\newtcolorbox{promptboxx}[1]{%
  enhanced,
  breakable,
  colback=white,
  colframe=blue!35,
  boxrule=0.5pt,
  arc=2pt,
  left=4pt,right=4pt,top=4pt,bottom=4pt,
  title=\textbf{#1},
  fonttitle=\small,
}

\usepackage{balance}         
\usepackage{subcaption}

\usepackage{graphicx}

\title{MAPGD: Multi-Agent Prompt Gradient Descent for Collaborative Prompt Optimization}

\author{
  \textbf{Yichen Han\textsuperscript{1,$\dagger$}\thanks{$\dagger$~Equal contribution. \ *~Corresponding authors.}},
  \textbf{Yuhang Han\textsuperscript{2,$\dagger$}},
  \textbf{Siteng Huang\textsuperscript{3}},
  \textbf{Guanyu Liu\textsuperscript{4}},
  \textbf{Zhengpeng Zhou\textsuperscript{2}},
  \textbf{Bojun Liu\textsuperscript{5}},
\\
  \textbf{Yujia Zhang\textsuperscript{7}},
  \textbf{Isaac N. Shi\textsuperscript{6}},
  \textbf{Lewei He\textsuperscript{1,*}},
  \textbf{Tianyu Shi\textsuperscript{8,*}}
\\[0.3em]
  \textsuperscript{1}South China Normal University;
  \textsuperscript{2}Shanghai Jiao Tong University;
  \textsuperscript{3}Zhejiang University;
  \textsuperscript{4}University of Macau;
\\
  \textsuperscript{5}University of Sydney;
  \textsuperscript{6}Silicon Sapiens LLC;
  \textsuperscript{7}University of Alberta;
  \textsuperscript{8}University of Toronto
\\[0.6em]
\href{mailto:helewei@m.scnu.edu.cn}{\texttt{helewei@m.scnu.edu.cn}}\quad
\href{mailto:tys@cs.toronto.edu}{\texttt{tys@cs.toronto.edu}}
}

\begin{document}
\maketitle
\begin{abstract}
Prompt engineering is crucial for fully leveraging large language models, yet existing prompt optimization methods often rely on a single refinement trajectory, leading to instability, conflicting update signals, and inefficient use of query budgets. We propose \textbf{M}ulti-\textbf{A}gent \textbf{P}rompt \textbf{G}radient \textbf{D}escent (MAPGD), a gradient-inspired framework that coordinates multiple complementary prompt editing signals for robust and interpretable optimization. MAPGD decomposes prompt refinement into orthogonal dimensions (e.g., instruction clarity, example selection, format, and style) and aggregates textual pseudo-gradients via semantic embedding, conflict-aware clustering, and adaptive fusion. To improve robustness, we introduce HCGC, which enforces angular separation between conflicting directions, and CAAW, which dynamically calibrates editing contributions based on validation feedback. Experiments on diverse classification and reasoning benchmarks show that MAPGD achieves consistent gains in accuracy over strong baselines, while ablation results indicate that coordinated gradient fusion and adaptive weighting are critical to stable optimization. Together, these findings suggest that MAPGD offers a robust and interpretable framework for automatic prompt optimization.

\end{abstract}

\section{Introduction}
\label{sec:introduction}

Large Language Models (LLMs), trained on vast web-scale corpora, have demonstrated remarkable generalization capabilities across a wide range of natural language processing (NLP) tasks, including classification, reasoning, and generation \cite{achiam2023gpt, bubeck2023sparks}. A critical factor underlying this performance is the design of prompts, which serve as the primary interface for steering model behavior. Despite their importance, prompts are still predominantly crafted through manual trial-and-error, requiring substantial human effort \cite{jiang2022promptmaker} and domain expertise \cite{reynolds2021prompt, zamfirescu2023johnny}. This reliance on manual design limits scalability and motivates the development of automatic or semi-automatic prompt optimization methods that are both effective and interpretable.

Existing approaches to prompt optimization largely fall into two categories. The first line of work approximates gradient-based optimization through auxiliary models or differentiable prompt representations \cite{qin2021learning, deng2022rlprompt}. While theoretically appealing, these methods often assume access to internal model states or parameters \cite{shin2020autoprompt, lester2021power}, making them impractical in real-world, API-only settings. The second line of work formulates prompt optimization as a discrete search or reinforcement learning problem, using feedback-driven edits or Monte Carlo exploration \cite{zhang2022tempera, zhou2022large}. Although effective in certain cases, these methods can suffer from high computational cost, unstable optimization trajectories, and limited interpretability. Collectively, these limitations highlight the need for more robust, scalable, and transparent prompt optimization frameworks.

To bridge this gap, Pryzant et al.~\cite{pryzant2023automatic} proposed ProTeGi, a non-parametric framework that reinterprets prompt optimization as gradient descent in natural language space. ProTeGi generates textual pseudo-gradients from model errors and iteratively refines prompts using beam search and bandit-based selection, achieving strong empirical performance. However, ProTeGi relies on a single-agent optimization framework that aggregates heterogeneous refinement signals into a unified trajectory. Such entanglement may give rise to implicit objective conflicts, thereby impairing optimization stability and convergence quality.

A natural extension is to decompose prompt optimization into multiple specialized perspectives. Multi-agent formulations enable parallel exploration of heterogeneous refinement dimensions, increasing signal diversity and reducing premature convergence. However, introducing multiple agents also makes semantic conflicts inevitable: different agents may propose mutually incompatible updates, such as simultaneously expanding illustrative examples while enforcing concise instructions. Without explicit coordination mechanisms, such conflicts can destabilize optimization and undermine the benefits of multi-agent collaboration.

\begin{figure*}[t]
\centering
\includegraphics[width=\textwidth]{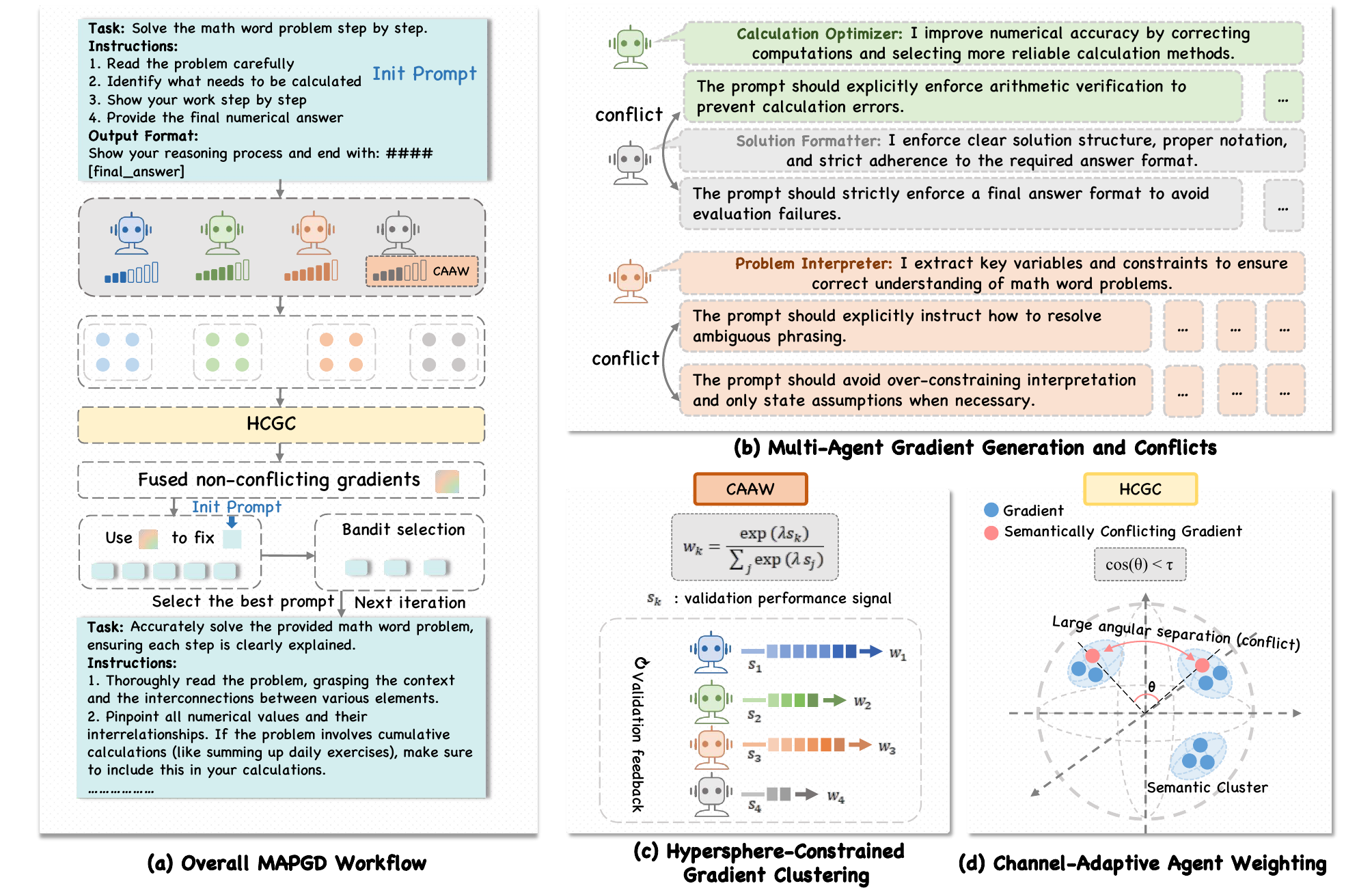}
\caption{Overview of \textbf{MAPGD}. MAPGD performs prompt optimization through collaborative multi-agent pseudo-gradients. Agent-generated gradients are embedded and clustered via HCGC to resolve semantic conflicts, and adaptively weighted by CAAW according to validation performance. The resulting fused gradients iteratively refine the prompt through candidate generation and selection.}
\label{fig:overview of mapgd}  
\end{figure*}

In this work, we propose MAPGD, a unified framework that addresses these challenges by integrating multi-agent specialization with principled coordination mechanisms (Figure~\ref{fig:overview of mapgd}). As illustrated in Figure~\ref{fig:overview of mapgd}(a), MAPGD decomposes prompt optimization into multiple orthogonal refinement dimensions and assigns each to a specialized agent. Each agent independently analyzes model errors and produces textual pseudo-gradients that capture targeted improvement directions.

However, pseudo-gradients produced by different agents may encode semantically conflicting refinement intents. To explicitly disentangle such conflicts, MAPGD proposes Hypersphere-Constrained Gradient Clustering (HCGC), which embeds agent-generated gradients into a shared semantic space and separates incompatible optimization directions via angularly constrained clustering. This design promotes semantic consistency within clusters while preventing destructive interference across conflicting refinement signals.

While HCGC addresses conflicts at the directional level, agent contributions within the same semantic cluster may still vary in reliability. MAPGD therefore incorporates CAAW, which dynamically reweights agent channels based on validation performance. By emphasizing consistently effective signals and attenuating noisy or unstable ones, CAAW enables robust aggregation of non-conflicting gradients, leading to stable and effective prompt refinement.

In summary, our contributions are three-fold.

(1) We propose MAPGD, a multi-agent pseudo-gradient framework for stable and interpretable prompt optimization.

(2) We introduce HCGC and CAAW to disentangle conflicting refinement signals and adaptively aggregate reliable agent contributions.

(3) We establish theoretical convergence guarantees, showing that MAPGD preserves classical gradient descent behavior with almost sure convergence to a local optimum at rate $\mathcal{O}(1/\sqrt{T})$.

\section{Related Work}
\label{sec:related_work}

\paragraph{Prompt Learning and Optimization.}
Early prompt engineering relied on manual design, which is labor-intensive and difficult to scale. Automated approaches have since been explored, including reinforcement learning \cite{deng2022rlprompt}, evolutionary search \cite{fernando2023promptbreeder}, and Bayesian optimization \cite{sahoo2025systematicsurveypromptengineering}. Continuous prompt tuning methods, such as prefix tuning \cite{Li2021PrefixTuningOC} and soft prompts \cite{lester2021power}, optimize prompts in embedding space but often sacrifice interpretability and require access to model internals. MAPGD operates in natural language space, preserving interpretability while enabling feedback-driven optimization.

\paragraph{Gradient-Inspired Prompt Optimization.}
Recent work has framed prompt optimization as a gradient-like process in discrete language space. ProTeGi \cite{pryzant2023automatic} generates natural language pseudo-gradients from model errors and iteratively refines prompts via beam search and bandit-based selection. However, as a single-agent method, ProTeGi entangles heterogeneous refinement objectives and lacks explicit mechanisms for resolving conflicting optimization signals, which can lead to unstable or suboptimal convergence. This limitation motivates approaches that can disentangle and coordinate diverse refinement directions.

\paragraph{Multi-Agent Collaboration.}
Multi-agent formulations enable parallel exploration of diverse perspectives and have been widely studied in reinforcement learning and distributed AI. In NLP, multi-agent debate \cite{liang2023mad, du2023improving} and collaborative generation \cite{hong2024metagptmetaprogrammingmultiagent} demonstrate the benefits of role specialization. However, increased diversity makes semantic conflicts inevitable, as different agents may propose incompatible updates. MAPGD adopts a multi-agent design with specialized roles, while explicitly addressing the resulting coordination and conflict resolution challenges.

\paragraph{Geometric and Adaptive Coordination.}
Resolving semantic conflicts requires structured inductive biases beyond heuristic fusion. In representation learning, contrastive objectives and angular margin constraints enforce intra-cluster compactness and inter-cluster separation \cite{wang2020understanding, liu2017sphereface, wang2018cosface, deng2019arcface}. Chen et al.~\cite{chen2021large} further enhance large-margin contrastive learning via distance polarization regularization. Inspired by these advances, MAPGD introduces HCGC for geometry-aware gradient coordination, complemented by CAAW to adaptively calibrate agent contributions.

\paragraph{Concurrent Work.}
Concurrent studies such as MAPRO \cite{zhang2025maprorecastingmultiagentprompt} and MA-SAPO \cite{seo2025promptoptimizationretrievedreasoning} also explore multi-agent prompt optimization under different assumptions, including explicit interaction structures or external memory components. These settings are not directly aligned with our budgeted, task-level optimization protocol. A detailed comparison is provided in Appendix~\ref{app:concurrent_work}.

\textbf{In summary}, MAPGD unifies prompt optimization, gradient-inspired refinement, multi-agent collaboration, and geometry-aware coordination into an interpretable framework that explicitly addresses semantic conflicts.


\section{Methodology}
\label{sec:methodology}

\subsection{Framework Overview}
MAPGD formulates prompt optimization as a hybrid discrete--continuous gradient descent process operating directly in natural language space. Unlike continuous embedding-based approaches such as prefix tuning or soft prompt optimization \cite{Li2021PrefixTuningOC, lester2021power}, MAPGD preserves interpretability by explicitly manipulating textual prompts while still leveraging gradient-inspired feedback signals for iterative refinement. As illustrated in Figure~\ref{fig:overview of mapgd}, MAPGD frames prompt optimization as a collaborative multi-agent process, in which specialized agents focus on orthogonal refinement dimensions and produce textual pseudo-gradients that guide prompt updates.

Building on this formulation, MAPGD instantiates the optimization procedure through an iterative coordination loop. At each iteration, specialized agents independently examine the current prompt and available task feedback, and generate natural language pseudo-gradients that reflect distinct refinement perspectives. These gradients are embedded into a shared semantic space, where potential conflicts are explicitly identified and mitigated via geometry-aware clustering and adaptive fusion. The resulting fused gradients guide the generation of successor prompts, which are subsequently evaluated and filtered under a budgeted selection mechanism. Repeating this process yields an evolving optimization trajectory that systematically balances exploration across heterogeneous refinement directions with exploitation of effective updates.

Formally, the prompt optimization objective is defined as
\begin{equation}
F(p) = \mathbb{E}_{(x,y) \sim \mathcal{D}} \left[ \ell(M(x; p), y) \right],
\end{equation}
where $\ell(\cdot,\cdot)$ denotes a task-specific loss function and $M(x; p)$ represents the model output conditioned on prompt $p$. The optimization goal is
\begin{equation}
p^* = \arg\min_p F(p).
\end{equation}
Since gradients in this discrete textual space are not directly accessible, MAPGD constructs natural language pseudo-gradients,
\begin{equation}
\nabla F(p^{(t)}) \approx g^{(t)},
\end{equation}
which serve as semantic analogues of numerical gradients in classical optimization.

\subsection{Specialized Prompt Agents}
MAPGD employs multiple specialized agents, each responsible for a distinct and approximately orthogonal dimension of prompt refinement. Typical agent roles include instruction clarity, example selection, output formatting, and stylistic refinement, optionally complemented by a generic agent for broad exploratory updates. At iteration $t$, the resulting pseudo-gradient set is
\[
G^{(t)} = \{g_1^{(t)}, g_2^{(t)}, \dots, g_K^{(t)}\}.
\]
By decomposing prompt optimization into specialized perspectives, MAPGD enables parallel exploration of heterogeneous refinement directions, alleviating the limited signal diversity and premature convergence commonly observed in single-agent optimization.
\vspace{-2mm}
\subsection{Hypersphere-Constrained Gradient Clustering}
\label{hcgc}

Inspired by hyperspherical metric learning \cite{wang2018cosface, liu2017sphereface},
HCGC resolves semantic conflicts among heterogeneous pseudo-gradients by enforcing
geometry-aware separation on a normalized hypersphere. Instead of assuming all
agent proposals are mutually compatible, HCGC explicitly detects semantic
disagreement and clusters gradients into coherent refinement directions before fusion.

\paragraph{Gradient Embedding and Conflict Signal.}
Given agent-generated pseudo-gradients $G^{(t)}$, each $g_k^{(t)}$ is encoded into a
$d$-dimensional semantic vector $v_k^{(t)}=\phi(g_k^{(t)})$ and normalized onto the unit
hypersphere:
\begin{equation}
\hat{v}_k^{(t)}=\frac{v_k^{(t)}}{\|v_k^{(t)}\|}, \quad \hat{v}_k^{(t)}\in\mathbb{S}^{d-1}.
\end{equation}
Semantic similarity is measured by cosine similarity,
\begin{equation}
\mathrm{sim}(\hat{v}_i,\hat{v}_j)=\hat{v}_i^\top \hat{v}_j=\cos(\Delta(\hat{v}_i,\hat{v}_j)),
\end{equation}
where $\Delta(\cdot,\cdot)$ denotes the angular distance on the hypersphere.
To make ``conflict'' explicit and measurable, we define two gradients as being in
semantic conflict if their similarity falls below a threshold:
\begin{equation}
\mathrm{conflict}(\hat{v}_i,\hat{v}_j)\;\Leftrightarrow\; \hat{v}_i^\top \hat{v}_j < \theta_{\text{conflict}}.
\end{equation}
Intuitively, low cosine similarity indicates substantially different (and potentially
incompatible) refinement intents, motivating explicit separation in clustering.

\paragraph{Clustering and Adaptive Cluster Cardinality.}
We perform cosine K-means on the hypersphere to group semantically coherent directions.
Rather than fixing the number of clusters, we adopt a bounded adaptive strategy:
\begin{equation}
K^{(t)}=\min\big(|G^{(t)}|,\;K_{\max}\big),
\end{equation}
where $|G^{(t)}|$ is the number of gradients at iteration $t$ and $K_{\max}$ is an upper
bound. This prevents over-fragmentation when gradients are few, while allowing multiple
distinct directions when agent proposals are diverse.

\paragraph{Angular Margin Constraint.}
Initial clustering may still yield weakly separated clusters when conflicting gradients
coexist. HCGC therefore refines assignments with an angular margin constraint.
Let $u_i$ be the centroid of $\hat{v}_k$'s assigned cluster and $u_j$ any other centroid.
Define $\alpha=\Delta(\hat{v}_k,u_i)$ and $\beta=\Delta(\hat{v}_k,u_j)$. While standard
assignment requires $\alpha<\beta$, HCGC strengthens it by a margin scale $n\ge 1$:
\begin{equation}
n\cdot \alpha < \beta,
\end{equation}
equivalently,
\begin{equation}
\cos(n\cdot \alpha) > \cos(\beta).
\end{equation}
Gradients that violate the margin are reassigned to clusters where the constraint holds,
yielding more compact and better-separated semantic groups.

\paragraph{Fusion of Clustered Gradients.}
Finally, gradients within each cluster are fused into a single representative refinement
direction via LLM-guided synthesis. To avoid redundant updates, we apply a diversity
filter to discard fused gradients overly similar to those already selected, controlled by
$\theta_{\text{diversity}}$.

\begin{figure}[t]
\centering
\includegraphics[width=0.5\textwidth]{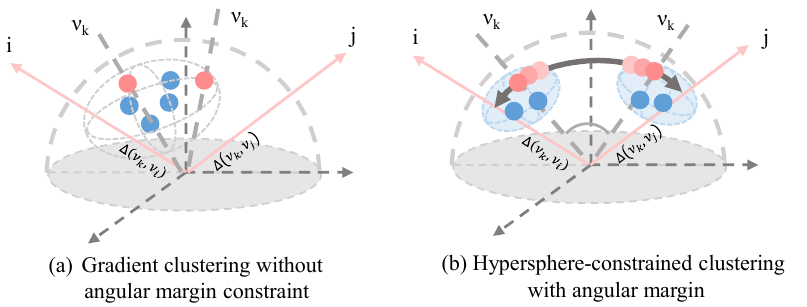}
\caption{Effect of hypersphere-constrained angular margin in HCGC.
(a) Clustering without angular margin constraints may entangle semantically conflicting pseudo-gradients, leading to ambiguous cluster boundaries.
(b) Enforcing an angular margin on the unit hypersphere improves intra-cluster compactness and inter-cluster separation, effectively isolating conflicting refinement directions.}
\label{fig:hcgc-margin}
\end{figure}
\vspace{-2mm}

\subsection{Channel-Adaptive Agent Weighting}
Although HCGC resolves structural conflicts between semantic gradient directions, gradients within the same cluster may still vary in reliability due to heterogeneous agent expertise and historical effectiveness. To account for this variability, we introduce CAAW, a reliability-aware coordination mechanism that adaptively calibrates agent contributions during gradient fusion. CAAW is inspired by channel-wise recalibration mechanisms in neural representations \cite{hu2018squeeze}, but operates at the level of agent-generated textual gradients.

\paragraph{Adaptive Weight Estimation.}
Let $s_k^{(t)}$ denote the validation performance gain attributed to gradient $g_k^{(t)}$ at iteration $t$, estimated from historical agent performance statistics. CAAW converts these scalar signals into normalized reliability weights via a softmax transformation:
\begin{equation}
w_k^{(t)} = \frac{\exp(\lambda s_k^{(t)})}{\sum_{j \in C} \exp(\lambda s_j^{(t)})},
\end{equation}
where $C$ denotes the set of gradients within the same semantic cluster and $\lambda$ controls the sharpness of the weighting distribution. This formulation assigns higher weights to agents that have consistently produced effective refinements, while down-weighting less reliable channels.

\paragraph{Weight-Conditioned LLM Fusion.}
Rather than explicitly performing an arithmetic weighted sum over textual gradients, CAAW injects the estimated weights into the LLM-driven fusion operator as conditioning signals. Concretely, the fusion operator
\begin{equation}
g_{\text{fused}}^{(t)} = \Psi \big( \{ g_k^{(t)} \}_{k \in C} \,\big|\, \{ w_k^{(t)} \}_{k \in C} \big)
\end{equation}
receives both the set of clustered gradients and their corresponding reliability weights, which are encoded in the fusion prompt to bias the synthesis process toward more trustworthy agent suggestions. In this way, agent weights influence the fused gradient implicitly through prompt-level emphasis, rather than through explicit numerical combination.

\paragraph{Stability Implications.}
By conditioning gradient fusion on adaptive reliability weights, CAAW suppresses noisy or redundant updates while amplifying consistently effective refinement channels. This weight-aware fusion mechanism reduces variance in the optimization trajectory and complements HCGC by addressing intra-cluster reliability differences, thereby promoting stable convergence toward semantically robust prompt candidates.

\vspace{-2mm}
\subsection{Candidate Generation and Selection}
Fused gradients generate successor prompts \(\{p'_1, \dots, p'_n\}\). MAPGD applies lightweight filtering and optional bandit-based evaluation \cite{audibert2010best} to keep the computation budgeted. Unlike prior work that depends heavily on bandit selection, MAPGD attains robustness and interpretability mainly via conflict-aware clustering and adaptive weighting across agents. Overall algorithm can be found in Algorithm~\ref{alg:mapgd-hcgc-caaw}.

\begin{algorithm}[t]
\caption{MAPGD with HCGC and CAAW}
\label{alg:mapgd-hcgc-caaw}
\begin{algorithmic}[1]
\Require Initial prompt $p_0$, train data $D_{\text{train}}$, dev data $D_{\text{dev}}$, agents $\{A_i\}_{i=1}^N$, iterations $R$, beam width $k$
\State Initialize best prompt $p_\star^{(0)} \gets p_0$, beam $B_0 \gets \{p_0\}$; initialize each agent prompt $A_i.p \gets p_0$
\For{$t=1$ to $R$}
   \State $M_t \gets \textsc{SampleMiniBatch}(D_{\text{train}}, b)$
   \State $\mathcal{G}_t \gets \textsc{GeneAgentGrads}(\{A_i\}, M_t)$ 
   \Comment{Alg.~\ref{alg:agent-grad}}
   \State $\tilde{\mathcal{G}}_t \gets \textsc{HCGC}(\mathcal{G}_t)$ \Comment{Alg.~\ref{alg:hcgc}}
   \State $F_t \gets \textsc{CAAW}(\tilde{\mathcal{G}}_t, D_{\text{dev}})$ \Comment{Alg.~\ref{alg:caaw}}
   \State $\mathcal{C}_t \gets \textsc{ExpandPrompts}(p_\star^{(t-1)}, F_t)$
   \State $B_t, p_\star^{(t)} \gets \textsc{BanditSelect}(\mathcal{C}_t, D_{\text{dev}}, k)$
   \State \textsc{SyncAgents}($\{A_i\}, p_\star^{(t)}$) \Comment{Alg.~\ref{alg:synchronize}}
   \If{\textsc{Converged}($p_\star^{(t)}$)} \textbf{break} \EndIf
\EndFor
\State \Return Best prompt over $\{p_\star^{(1)},\dots,p_\star^{(t)}\}$
\end{algorithmic}
\end{algorithm}

\vspace{-2mm}
\subsection{Theoretical Convergence}

We provide a theoretical analysis of MAPGD in Appendix~\ref{app:theory}. 
Here, we summarize the key assumptions and main implications to clarify how the proposed framework relates to classical stochastic optimization.

Specifically, our analysis relies on three standard conditions: 
(i) bounded second moments of agent-generated pseudo-gradients, 
(ii) smoothness (or Lipschitz continuity) of the empirical loss function in the semantic embedding space, 
(iii) unbiased sampling induced by the bandit-based candidate selection mechanism. 
Under these assumptions, MAPGD admits an interpretation as a stochastic approximation process and achieves a sublinear convergence rate of $\mathcal{O}(1/\sqrt{T})$ in both convex and non-convex settings.

Intuitively, the core components of MAPGD play complementary roles in ensuring stable optimization. 
HCGC aggregates semantically aligned refinement directions and mitigates conflicting updates, reducing variance in the fused pseudo-gradient. 
CAAW further emphasizes agents that have demonstrated consistent utility across iterations. 
Finally, bandit-based candidate selection maintains sufficient exploration while preventing systematic bias.
Together, these mechanisms ensure that the overall optimization dynamics remain stable and progressively improve prompt quality over iterations.


\vspace{-2mm}
\section{Experiments}
\vspace{-2mm}
Experiments evaluate MAPGD in terms of effectiveness, efficiency, and robustness.
For fair comparison, we reproduce ProTeGi under identical settings and further conduct ablations and token consumption analysis.
A real-world text-generation case study is also included (Appendix~\ref{app:casestudy}) to demonstrate applicability beyond benchmarks.

\vspace{-2mm}
\subsection{Experimental Details}
\label{sec:setup}
\vspace{-0.25em}

\noindent\textbf{Datasets.}
Following ProTeGi's protocol, we evaluate on four classification benchmarks
(Jailbreak, Ethos, LIAR, Sarcasm) and three arithmetic reasoning datasets
(GSM8K, AQUARAT, SVAMP). For each dataset, we sample 50/150 instances for
dev/test and report test F1 (classification) or accuracy (reasoning).
Dataset details are in Appendix~\ref{app:dataset_details}.
\vspace{-0.25em}

\noindent\textbf{Setup.}
We use the January 2023 \texttt{gpt-3.5-turbo} via Azure OpenAI unless stated
otherwise (temperature $0.0$ for classification). Results are averaged over
three runs; F1 is computed by max-pooling over the final prompt beam. All
methods share identical initial prompts, data, and random seeds, with a fixed
budget of $T{=}10$ and default hyperparameters. Unless stated otherwise, MAPGD follows Section~\ref{hcgc} with
$\phi{=}\texttt{all-MiniLM-L6-v2}$, $\theta_{\text{conflict}}{=}0.3$, clustering
threshold $0.7$, $\theta_{\text{diversity}}{=}0.7$, and UCB selection with 80
evaluations; $\lambda{=}1$ (Appendix~\ref{lambda}). All experiments in
Section~\ref{sec:results} use these modules unless ablated.
\vspace{-0.25em}

\noindent\textbf{Baselines.}
We primarily compare against ProTeGi~\cite{pryzant2023automatic} under identical
query budgets and initialization, and additionally include MC search, RL-style
edit methods, AutoGPT, and three reasoning baselines (InsZero, Instinct,
PromptWizard). Full descriptions are in Appendix~\ref{app:baseline_details}.

\vspace{-2mm}
\subsection{Experimental Results}
\vspace{-1mm}
\label{sec:results}

\begin{figure}[t]
\centering
\includegraphics[width=\columnwidth]{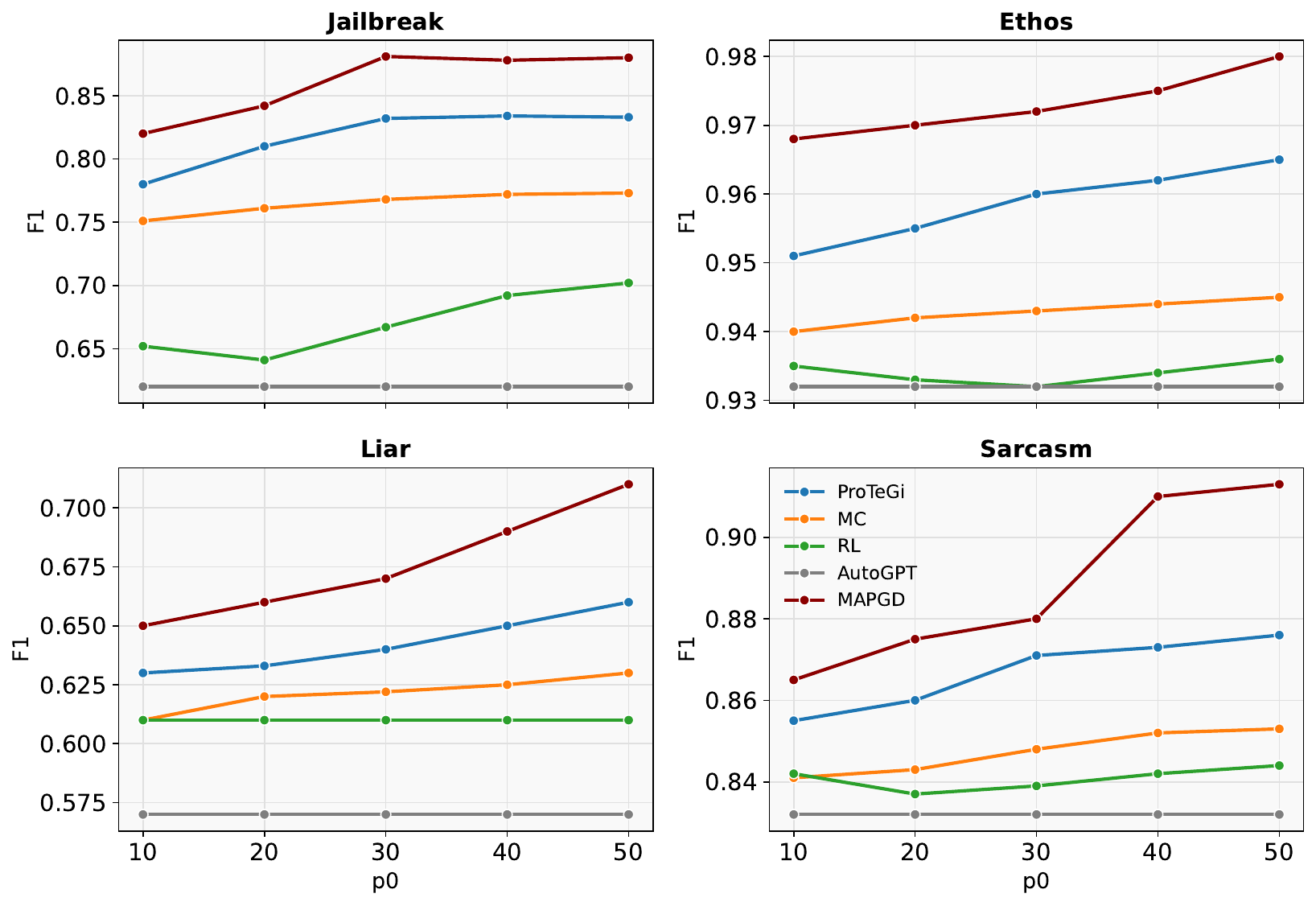}
\caption{Test performance (F1 score) vs. API query budget per prompt candidate across four benchmark tasks (evaluated up to 50 queries for fair comparison with ProTeGi).}
\label{fig:overall-results}
\end{figure}

\textbf{Classification benchmarks.}
Figure~\ref{fig:overall-results} reports test F1 under different per-candidate evaluation budgets (up to 50 queries, following ProTeGi for fairness). MAPGD consistently achieves the best performance across all four datasets, with most gains emerging within the first 30--40 evaluations. The improvements are most pronounced on conflict-prone tasks, suggesting that \emph{explicit conflict disentanglement and adaptive coordination} are critical when refinement signals disagree: on \textbf{Jailbreak}, MAPGD reaches \textbf{0.880} at 50 queries (vs.\ 0.833 for ProTeGi), and on \textbf{LIAR} it attains \textbf{0.710} (vs.\ 0.660). We attribute these gains to HCGC, which clusters compatible pseudo-gradients before fusion to reduce semantic interference, and CAAW, which down-weights unreliable agents to prevent noisy edits from dominating early iterations, thereby improving budget efficiency. On the more saturated \textbf{Ethos} dataset, MAPGD still improves to \textbf{0.980} (vs.\ 0.965), indicating that coordinated multi-agent feedback can extract residual refinements even near a performance ceiling. We follow the 50-query budget for strict comparability; extended-budget results are deferred to Appendix~\ref{app:budget-extension}.

\textbf{Arithmetic reasoning benchmarks.}
Table~\ref{tab:arithmetic-results} evaluates MAPGD on GSM8k, AQUARAT, and SVAMP, compared against InsZero, Instinct, and PromptWizard. MAPGD achieves consistent gains across all three datasets, most notably on \textbf{GSM8k} (\textbf{93.5}\%), improving over the strongest baseline by +3.5 points. We hypothesize that the benefit on reasoning stems from the same coordination principle: specialized agents propose complementary edits (e.g., step-by-step constraints, format strictness, and exemplar guidance), while HCGC/CAAW suppress incompatible or low-yield directions, leading to more stable chain-of-thought prompting. MAPGD also improves the best baseline on \textbf{AQUARAT} (60.3\% vs.\ 58.2\%) and \textbf{SVAMP} (84.1\% vs.\ 82.3\%), confirming transfer beyond classification.


\begin{table}[!t]
  \centering
  \tablestyle{5pt}{1.0}
  \setlength\tabcolsep{3.5pt}
  \begin{tabular}{c|ccc}
  \textbf{Method} & \textbf{GSM8k} & \textbf{AQUARAT} & \textbf{SVAMP} \\
  \midrule
  InsZero      & 74.2 & 54.3 & 79.5 \\
  Instinct     & 74.5 & 54.7 & 81.0 \\
  PromptWizard & 90.0 & 58.2 & 82.3 \\
  \rowcolor[rgb]{.949,.949,.949} MAPGD & \textbf{93.5} & \textbf{60.3} & \textbf{84.1} \\
\end{tabular}
  \caption{Arithmetic reasoning accuracy (\%) on GSM8k, AQUARAT, and SVAMP. Best results in bold.}
  \label{tab:arithmetic-results}
\end{table}%


\vspace{-2mm}
\subsection{Ablation Study}
To isolate the sources of MAPGD’s gains, we conduct a component-wise ablation by selectively disabling HCGC and CAAW. Table~\ref{tab:ablation-components} reports four settings (I--IV), where Setting~I removes both modules and serves as a minimal baseline. Enabling either module leads to consistent improvements. Introducing HCGC (III vs.\ I) yields systematic gains, confirming that geometry-aware clustering prior to fusion is essential for reducing semantic interference among pseudo-gradients. CAAW alone (II vs.\ I) also improves performance, indicating that uniform averaging is suboptimal under varying agent reliability; adaptive reweighting based on validation feedback enhances robustness. The combined setting (IV) consistently outperforms either component alone, suggesting strong complementarity: HCGC resolves directional incompatibility via conflict disentanglement, whereas CAAW addresses reliability mismatch through adaptive credit assignment. Gains are smaller on saturated datasets (e.g., Ethos), consistent with ceiling effects, while conflict-prone tasks benefit more from structured coordination.

\begin{table}[t]
\centering
\caption{Ablation study on core MAPGD components. $\checkmark$ indicates the component is enabled. Results are F1 score (mean over 3 runs).}
\label{tab:ablation-components}
\resizebox{\linewidth}{!}{
\begin{tabular}{c cc cccc}
\toprule
\textbf{Setting} 
& \textbf{HCGC} 
& \textbf{CAAW} 
& \textbf{Jailbreak} 
& \textbf{Ethos} 
& \textbf{LIAR} 
& \textbf{Sarcasm} \\
\midrule
I   & $\times$     & $\times$     & 0.82 & 0.91 & 0.62 & 0.83 \\
II  & $\times$     & $\checkmark$ & 0.84 & 0.94 & 0.62 & 0.87 \\
III & $\checkmark$ & $\times$     & 0.85 & 0.95 & 0.63 & 0.88 \\
IV  & $\checkmark$ & $\checkmark$ & \textbf{0.88} & \textbf{0.98} & \textbf{0.65} & \textbf{0.91} \\
\bottomrule
\end{tabular}
}
\end{table}

\vspace{-2mm}
\subsection{Robustness to Dataset Sampling Size}
\label{sec:robustness-sampling}

Following the standard ProTeGi evaluation protocol, our main results use 50 development and 150 test examples per dataset to ensure direct comparability. To verify that MAPGD's gains are not an artifact of small evaluation sets, we additionally evaluate larger subsamples on three representative classification benchmarks.

Table~\ref{tab:sampling-robustness} reports results under three sampling regimes (50/150, 100/300, and 200/500). MAPGD consistently outperforms ProTeGi and MC across all datasets and sampling sizes, indicating that its improvements persist as the evaluation set grows.

\begin{table}[!t]
  \centering
  \tablestyle{5pt}{1.0}
  \setlength\tabcolsep{3.5pt}
  \begin{tabular}{l|lccc}
    \textbf{Dataset} & \textbf{Method} & \textbf{50/150} & \textbf{100/300} & \textbf{200/500} \\
    \midrule
    Jailbreak & ProTeGi & 0.82 & 0.83 & 0.81 \\
             & MC      & 0.76 & 0.77 & 0.75 \\
    \rowcolor[rgb]{.949,.949,.949} 
             & \textbf{MAPGD} & \textbf{0.86} & \textbf{0.86} & \textbf{0.84} \\
    \midrule
    Ethos    & ProTeGi & 0.93 & 0.94 & 0.94 \\
             & MC      & 0.89 & 0.90 & 0.90 \\
    \rowcolor[rgb]{.949,.949,.949} 
             & \textbf{MAPGD} & \textbf{0.98} & \textbf{0.97} & \textbf{0.97} \\
    \midrule
    LIAR     & ProTeGi & 0.65 & 0.62 & 0.63 \\
             & MC      & 0.62 & 0.63 & 0.63 \\
    \rowcolor[rgb]{.949,.949,.949} 
             & \textbf{MAPGD} & \textbf{0.71} & \textbf{0.71} & \textbf{0.69} \\
  \end{tabular}
  \caption{Robustness to dataset sampling size (F1 score) under three dev/test splits.}
  \label{tab:sampling-robustness}
\end{table}%
\vspace{-2mm}
\subsection{Token Consumption Analysis}
Token efficiency is evaluated under the setup of Section~\ref{sec:setup} by counting both input and output tokens per API call. We report total tokens, number of calls, and performance-per-token.
As shown in Table~\ref{tab:token-consumption}, MAPGD attains higher F1 with fewer tokens and calls than ProTeGi, reducing per-iteration token cost by $\sim$8\% and improving performance-per-token by $>$10\%, suggesting reduced redundant queries via HCGC and CAAW.



\begin{table}[!t]
  \centering
  \tablestyle{5pt}{1.0}
  \setlength\tabcolsep{3.5pt}
  \begin{tabular}{c|cc}
    \textbf{Metric} & \textbf{ProTeGi} & \textbf{MAPGD (full)} \\
    \midrule
    Total Tokens & 256k & \cellcolor[rgb]{.949,.949,.949}\textbf{236k} \\
    Calls        & 962  & \cellcolor[rgb]{.949,.949,.949}\textbf{643} \\
    Avg F1       & 0.83 & \cellcolor[rgb]{.949,.949,.949}\textbf{0.87} \\
    F1 / Token ($\times 10^{-6}$) & 3.24 & \cellcolor[rgb]{.949,.949,.949}\textbf{3.69} \\
  \end{tabular}
  \caption{Token consumption comparison between ProTeGi and MAPGD (averaged over 3 runs). MAPGD achieves lower token usage and higher efficiency.}
  \label{tab:token-consumption}
\end{table}%

\vspace{-2mm}
\subsection{Computational Overhead Analysis}

We compare runtime overhead against ProTeGi under identical settings. MAPGD shares evaluation, prompt expansion, and bandit selection with ProTeGi, but adds multi-agent gradient generation and embedding-based clustering.
As shown in Table~\ref{tab:runtime-breakdown}, the extra overhead mainly comes
from multi-agent gradient generation, which scales linearly with the number of
agents. Embedding and HCGC incur an almost constant cost, and with four agents
the overall overhead remains a small fraction of total runtime.



\begin{table}[!t]
  \centering
  \tablestyle{5pt}{1.0}
  \setlength\tabcolsep{3.5pt}
  \begin{tabular}{l|cc}
    \textbf{Component} & \textbf{ProTeGi (s)} & \textbf{MAPGD (s)} \\
    \midrule
    Gradient generation & 10.2  & 37.9 \\
    Embedding + HCGC    & --   & 24.4 \\
    Prompt expansion    & 33.7 & 28.7 \\
    Bandit selection    & 30.3 & 26.5 \\
    Other               & 85.2 & 84.0 \\
    \midrule
    \rowcolor[rgb]{.949,.949,.949} \textbf{Total} & \textbf{159.4} & \textbf{201.5} \\
  \end{tabular}
  \caption{Runtime breakdown (in seconds) comparing ProTeGi and MAPGD.}
  \label{tab:runtime-breakdown}
\end{table}%
\vspace{-2mm}

\subsection{Agent Number Sensitivity}
\begin{figure}[t]
\centering
\includegraphics[width=0.45\textwidth]{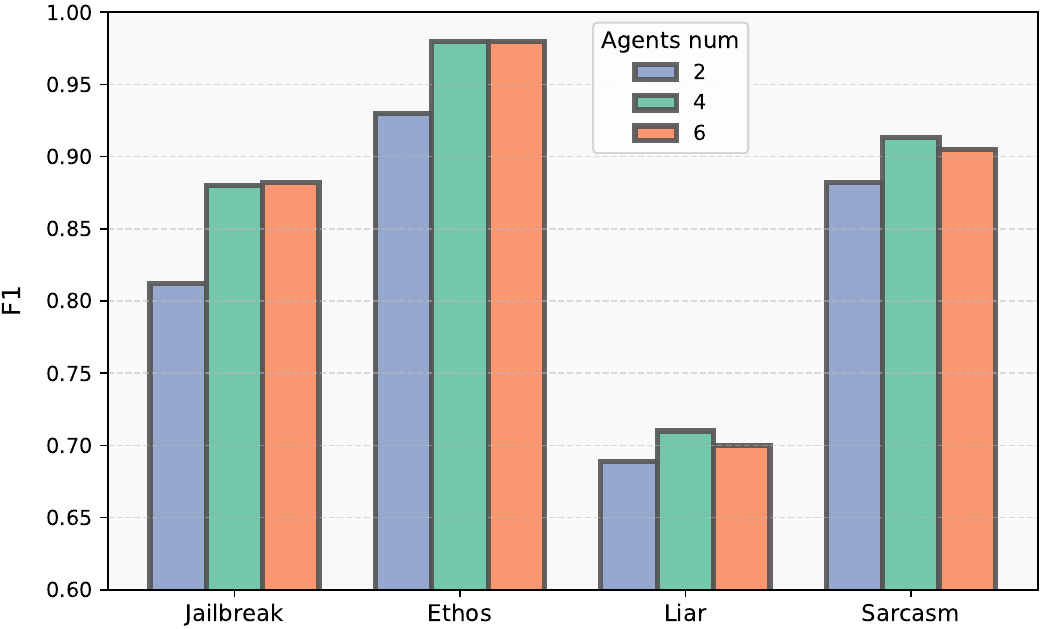}
\caption{Agent number sensitivity analysis. Increasing agents from 2 to 4 yields clear improvements, while further increasing to 6 brings diminishing returns.}
\label{fig:agent-sensitivity}
\end{figure}

To motivate the use of four specialized agents, we vary the number of agents $N\in\{2,4,6\}$ under the setup of Section~\ref{sec:setup}. For $N=2$, only the Instruction Specialist and Example Curator are used; for $N=6$, two additional style/format agents are added.
As shown in Figure~\ref{fig:agent-sensitivity}, average F1 improves substantially from $N=2$ to $N=4$, while $N=6$ yields only marginal gains with higher token cost, indicating diminishing returns. Thus, $N=4$ offers the best effectiveness--efficiency trade-off.



\vspace{-2mm}
\subsection{Experimental Insights}
MAPGD consistently outperforms single-agent ProTeGi and other baselines, with the largest gains on LIAR and Jailbreak where refinement signals often conflict. Ablations show HCGC and CAAW are complementary: removing either hurts performance, and removing both approaches the single-agent regime. MAPGD is also more token-efficient than ProTeGi (fewer API calls, $>$10\% higher performance-per-token) and transfers well to arithmetic reasoning, surpassing strong baselines on GSM8k, AQUARAT, and SVAMP. Appendix~\ref{app:qualitative-case} provides a qualitative case study
illustrating how HCGC and CAAW resolve conflicting pseudo-gradients.


\vspace{-2mm}
\section{Conclusion}
\vspace{-1mm}
This work introduces MAPGD, a coordinated multi-agent framework for prompt optimization that integrates HCGC and CAAW to resolve conflicting refinement directions and reduce the impact of unreliable agents.
Across classification and arithmetic reasoning benchmarks, MAPGD consistently improves over strong baselines while lowering token consumption, and it admits a sublinear convergence guarantee of $O(1/\sqrt{T})$ under standard stochastic approximation assumptions (Appendix~\ref{app:theory}).
Overall, MAPGD provides an interpretable, robust, and resource-efficient approach to prompt optimization via structured multi-agent collaboration.





\vspace{-2mm}
\section*{Limitations}
\vspace{-1mm}
MAPGD relies on semantic embeddings to detect conflicts and cluster pseudo-gradients; under domain shift or highly specialized jargon, similarity estimates may become unreliable and degrade coordination quality. The method also depends on reasonably informative agent feedback: if multiple agents are systematically misaligned, adaptive weighting may not fully suppress harmful edits. Finally, compared to single-agent methods, MAPGD introduces extra overhead from multi-agent generation and coordination (embedding/clustering), which may limit deployment under strict latency or budget constraints.


\bibliography{custom}

\appendix

\section{Datasets Details}
\label{app:dataset_details}
To ensure comparability with ProTeGi, we evaluate MAPGD on the same four benchmark NLP classification tasks that span diverse domains and languages:  

\begin{itemize}
    \item \textbf{Jailbreak} \cite{shen2024anything}: a novel task to determine whether a user input to an LLM continuation API constitutes a jailbreak attempt. Jailbreak attacks are defined as user strategies that aim to make the model violate its own safety policies, such as generating harmful content or exposing hidden instructions. The dataset contains 1306 multilingual examples with human-annotated jailbreak labels.  
    \item \textbf{Ethos} \cite{mollas2022ethos}: an English hate speech detection dataset with 997 online comments annotated as hateful or non-hateful.  
    \item \textbf{LIAR} \cite{wang2021new}: a large-scale dataset for fake news detection, consisting of 4,000 short statements labeled with ground-truth veracity, along with context information.  
    \item \textbf{Sarcasm} \cite{farha2020arabic}: an Arabic sarcasm detection dataset with 10,000 online comments, labeled for the presence or absence of sarcasm.  
\end{itemize}

Beyond classification, we further evaluate on three arithmetic reasoning datasets that require multi-step logical inference and symbolic manipulation:  

\begin{itemize}
    \item \textbf{GSM8k} \cite{cobbe2021training}: a widely used benchmark of 8,500 grade-school math problems that require step-by-step reasoning.  
    \item \textbf{AQUARAT} \cite{ling2017program}: a dataset of algebraic word problems designed to test symbolic reasoning and program induction.  
    \item \textbf{SVAMP} \cite{patel2021nlp}: a benchmark focusing on simple arithmetic word problems with diverse linguistic variations, serving as a testbed for robustness to paraphrasing.  
\end{itemize}

In addition, we include a real-world text generation case study in the Appendix to highlight MAPGD’s applicability beyond controlled benchmarks.

\section{Baselines Details}
\label{app:baseline_details}
To evaluate the effectiveness of MAPGD, we compare it against a set of non-parametric prompt optimization methods, following the setup of ProTeGi \cite{pryzant2023automatic}, and additionally include ProTeGi itself as a strong baseline. Specifically, we consider:  

\begin{itemize}
    \item \textbf{ProTeGi}: the original prompt gradient descent framework, where a single agent iteratively generates pseudo-gradients and candidate prompts, with bandit-based selection. This serves as our primary baseline for comparison.  
    \item \textbf{Monte-Carlo (MC)} \cite{zhou2022large}: an iterative but directionless Monte Carlo search over the prompt space. For fairness, we match the number of samples per candidate to the successors generated by MAPGD.  
    \item \textbf{Reinforcement Learning (RL)} Recent approaches such as GrIPS \cite{prasad2022grips} and TEMPERA \cite{zhang2022tempera} formulate prompt optimization as a reinforcement learning problem. In these methods, the prompt text is first segmented into phrases, and the search space is explored via phrase-level edit operations, including addition, paraphrasing, swapping, and deletion.
    \item \textbf{AutoGPT}: an open-source autonomous agent system that improves prompts through self-directed feedback loops. We configure AutoGPT with the same number of examples and errors as MAPGD, running for the same number of optimization steps. 
\end{itemize}

For arithmetic reasoning, we additionally compare against three competitive baselines, all reported with GPT-3.5-Turbo in a zero-shot setting:  

\begin{itemize}
    \item \textbf{InsZero} \cite{chen2023instructzeroefficientinstructionoptimization}: a zero-shot prompting method with instruction tuning.  
    \item \textbf{Instinct} \cite{lin2024useinstinctinstructionoptimization}: an instruction-following baseline that improves prompting robustness.  
    \item \textbf{PromptWizard} \cite{agarwal2024promptwizardtaskawarepromptoptimization}: a state-of-the-art prompt synthesis framework that iteratively enriches prompts with intermediate reasoning steps and examples.  
\end{itemize}

This collection of baselines allows us to evaluate MAPGD not only against single-agent gradient-based optimization, but also against strong arithmetic reasoning methods optimized for chain-of-thought style prompting.

\section{Extended Evaluation Budgets Beyond 50 Queries}
\label{app:budget-extension}

To verify whether our conclusions depend on the 50-query budget, we further extend the evaluation budget to \textbf{60} and \textbf{70} queries on three representative datasets (Jailbreak, LIAR, and Sarcasm). We keep the same evaluation protocol and report results in Table~\ref{tab:budget-extension}. Overall, the \textbf{method ranking remains unchanged}, and the additional gains beyond 50 are \textbf{marginal}, suggesting that a 50-query budget is sufficient to capture MAPGD’s performance while preserving fairness with prior work (e.g., ProTeGi).

\begin{table}[t]
\centering
\small
\setlength{\tabcolsep}{6pt}
\begin{tabular}{l c c c c}
\toprule
Dataset & Budget & ProTeGi & MC & MAPGD \\
\midrule
Jailbreak & 50 & 0.833 & 0.773 & 0.880 \\
         & 60 & 0.836 & 0.773 & 0.885 \\
         & 70 & 0.842 & 0.773 & 0.887 \\
\midrule
LIAR     & 50 & 0.660 & 0.630 & 0.710 \\
         & 60 & 0.672 & 0.630 & 0.720 \\
         & 70 & 0.681 & 0.630 & 0.725 \\
\midrule
Sarcasm  & 50 & 0.876 & 0.853 & 0.913 \\
         & 60 & 0.879 & 0.853 & 0.918 \\
         & 70 & 0.880 & 0.853 & 0.920 \\
\bottomrule
\end{tabular}
\caption{Performance under extended evaluation budgets (60/70 queries). Gains beyond 50 queries are small and do not change the ranking among methods.}
\label{tab:budget-extension}
\end{table}

Across these datasets, MAPGD improves by only \textbf{+0.007} (Jailbreak), \textbf{+0.015} (LIAR), and \textbf{+0.007} (Sarcasm) from 50 to 70 queries, confirming a clear diminishing-return trend after approximately 30--40 evaluations, while leaving the main conclusions unchanged.

\section{CAAW Parameter Sensitivity}

\label{lambda}

The CAAW module uses a temperature parameter $\lambda$ to control the emphasis on historically effective agents during gradient fusion. To evaluate the robustness of MAPGD with respect to $\lambda$, we conducted experiments on two representative datasets: \textbf{LIAR}(binary) and \textbf{GSM8K} (mathematical reasoning).

We tested three values of $\lambda$: 0.5, 1, and 2. Table~\ref{tab:lambda_sensitivity} summarizes the test accuracy for each setting. The results indicate that moderate variations of $\lambda$ do not significantly affect the overall performance, supporting the choice of $\lambda=1$ in the main experiments.

\begin{table}[!t]
\centering
\caption{Test accuracy of MAPGD under different CAAW weighting parameter $\lambda$. Performance differences are minor, demonstrating robustness to $\lambda$.}
\label{tab:lambda_sensitivity}
\begin{tabular}{c|cc}
\hline
$\lambda$ & LIAR Accuracy & GSM8K Accuracy \\
\hline
0.5 & 0.709 & 0.927 \\
1   & 0.717 & 0.935 \\
2   & 0.711 & 0.933 \\
\hline
\end{tabular}
\end{table}

As shown in Table~\ref{tab:lambda_sensitivity}, the test accuracy exhibits only minor variations across the three $\lambda$ values, with a maximum difference of 0.008 on LIAR and 0.008 on GSM8K. These variations are small relative to the overall performance gains achieved by multi-agent gradient fusion, indicating that while $\lambda$ influences the sharpness of agent weighting, MAPGD's performance remains stable for reasonable choices around the default setting of $\lambda=1$.

\section{Comparison with Recent Concurrent Work}
\label{app:concurrent_work}

We are aware of two recent concurrent works that are closely related in spirit to prompt optimization with multi-agent components, namely MAPRO~\cite{zhang2025maprorecastingmultiagentprompt} and MA-SAPO~\cite{seo2025promptoptimizationretrievedreasoning}. While these works share a high-level motivation with MAPGD, we did not include them as baselines in our main experimental comparisons due to comparability constraints in problem setting and resource assumptions.

\paragraph{MAPRO~\cite{zhang2025maprorecastingmultiagentprompt}.}
MAPRO studies prompt optimization \emph{within an explicit multi-agent system (MAS)}: it jointly configures role prompts for multiple agents and leverages the MAS interaction topology (e.g., DAG-structured communication) together with node-/edge-level reward modeling and global inference to select prompt combinations. In contrast, MAPGD targets \emph{task-level prompt optimization} under a fixed query budget, where specialized agents propose pseudo-gradients that are subsequently coordinated (via HCGC and CAAW) to iteratively refine a \emph{single} task prompt. Because MAPRO's formulation depends on an explicit MAS graph, per-agent prompts, and interaction-specific rewards, directly porting it to our benchmark protocol would require redesigning the environment and supervision signals, making a head-to-head comparison not directly aligned with our evaluation setting.

\paragraph{MA-SAPO~\cite{seo2025promptoptimizationretrievedreasoning}.}
MA-SAPO optimizes prompts through \emph{retrieved reasoning assets}: it constructs a searchable library of semi-structured ``reasoning cards'' from labeled or scored prompt--response data, and then performs analysis--refinement guided by top-$k$ retrieved assets at test time. This introduces an additional external memory/resource assumption (i.e., an offline asset corpus and its construction pipeline) that is orthogonal to our training-free, black-box optimization protocol. In our evaluation, we intentionally restrict methods to operate under the same query budget without relying on extra annotated corpora or a separately curated retrieval memory. Under these constraints, implementing MA-SAPO would either (i) require additional data collection/annotation to build assets, or (ii) reuse task-specific optimization traces as assets, which changes the method definition and complicates fairness.

\paragraph{Summary and future work.}
Overall, MAPRO and MA-SAPO represent complementary directions—MAS-level joint role-prompt configuration and retrieval-augmented prompt rewriting—whereas MAPGD focuses on multi-agent pseudo-gradient coordination for budgeted task-level prompt refinement. Integrating these paradigms into a unified and strictly comparable evaluation (e.g., extending MAPGD with optional retrieval memories or evaluating MAPRO-style joint prompting on explicit MAS benchmarks) is an important direction for future work.


\section{Complexity and Parallelism}
\label{Complexity and Parallelism}
The computational bottleneck of MAPGD with HCGC and CAAW (see Algorithm~\ref{alg:mapgd-hcgc-caaw}) lies in LLM calls for gradient generation, cluster fusion, and prompt expansion, whereas embedding, clustering, similarity checks, and adaptive weighting involve comparatively lightweight vector operations. Multi-agent parallelization significantly reduces wall time, making the framework scalable for practical deployment. In the following, we analyze the per-iteration computational and memory costs, providing a detailed breakdown of the dominant operations and their respective complexities.

\paragraph{Notation.}  
Let $N$ be the number of agents, $m$ the number of reasons per agent, giving $G = N m$ atomic gradients.  
Embedding dimension $d$, clusters $K \le G$, fused gradients $|\tilde{\mathcal{G}}|$, expansion variants per gradient $s$, MC paraphrases per variant $n_{\text{mc}}$, candidate prompts $|\mathcal{C}|$, beam width $k$, bandit rounds $T_b$ with $K_{\text{eval}}$ arms and dev mini-batch size $b$.  

\paragraph{Stage costs (time).}
\begin{enumerate}
\item \textbf{Agent gradient generation:} 
Each of the $N$ agents generates $m$ gradients using LLMs, giving $O(N)$ LLM calls. 
Parallelization reduces wall time to $\approx \max t_{\text{LLM}}$.

\item \textbf{HCGC embedding + conflict detection:} 
Embedding each gradient costs $O(G d)$; computing pairwise angular conflicts costs $O(G^2 d)$.  
For small $G$ (e.g., $G \sim 16$), this is negligible compared with LLM calls.

\item \textbf{HCGC clustering:} 
KMeans clustering over $G$ gradients into $K$ clusters with $I$ iterations costs $O(G K I d)$, again minor due to small $G$.

\item \textbf{HCGC fusion (LLM):} 
Clusters with multiple gradients require one LLM call per cluster to fuse into a coherent gradient.  
Total LLM calls up to $O(K_{\text{merge}})$, parallelizable across clusters.

\item \textbf{CAAW weighting and fusion:} 
Per-cluster adaptive weighting is $O(G)$ vector operations, negligible compared to LLM fusion.

\item \textbf{Prompt expansion + MC paraphrasing:} 
Applying each fused gradient: $O(|\tilde{\mathcal{G}}|)$ gradient applications.  
MC paraphrasing for each variant: $O(|\tilde{\mathcal{G}}| s n_{\text{mc}})$ LLM calls.

\item \textbf{Diversity filtering:} 
Computing embeddings: $O(|\mathcal{C}| d)$; naive pairwise similarities: $O(|\mathcal{C}|^2)$.  
For tens of candidate prompts, this remains negligible.

\item \textbf{Bandit evaluation:} 
Probing $K_{\text{eval}}$ arms over mini-batches of size $b$ for $T_b$ rounds: $O(K_{\text{eval}} b T_b)$, significantly cheaper than exhaustive evaluation $O(|\mathcal{C}|\,|D_{\text{dev}}|)$.
\end{enumerate}

\paragraph{Space complexity.}  
Text storage: $O(|\mathcal{C}| L_{\text{avg}})$.  
Embeddings: $O((G + |\mathcal{C}|) d)$, typically a few MB.  
Optional caches (unique prompts or intermediate LLM outputs) scale linearly with the number of prompts and clusters.


\begin{algorithm}[t]
\caption{Multi-Agent Textual Gradient Generation}
\label{alg:agent-grad}
\begin{algorithmic}[1]
\Require Agents $\{A_i\}$, mini-batch $M_t$, task $T$, predictor $\Pi$, per-agent error cap $e$, feedback count $m$
\ForAll{agent $A_i$ in parallel}
    \State $(\hat{y}, y)$ pairs $\gets T.\textsc{InferAndLabel}(A_i.p, M_t, \Pi)$
    \State $E_i \gets \textsc{SelectErrors}(\hat{y}, y, e)$
    \If{$|E_i|=0$} $E_i \gets \textsc{DiverseSamples}(M_t, e)$ \EndIf
    \State $raw_i \gets \textsc{LLMGradientPrompt}(A_i.role, A_i.p, E_i, m)$
    \State $g_i \gets \textsc{ParseGradientBlocks}(raw_i)$ \Comment{Split by delimiters}
\EndFor
\State \Return $\mathcal{G}_t = \{(A_i.role, g_i)\}_{i=1}^N$
\end{algorithmic}
\end{algorithm}

\begin{algorithm}[t]
\caption{Hypersphere-Constrained Gradient Clustering}
\label{alg:hcgc}
\begin{algorithmic}[1]
\Require Agent gradients $\mathcal{G}_t$ 
\State Embed and normalize: $\hat{v}_k = \phi(g_k)/\|\phi(g_k)\|$
\State Detect conflicts: $\mathcal{C}_{\text{conf}} \gets \{(i,j): \cos^{-1}(\hat{v}_i^\top \hat{v}_j) > \theta\}$
\State Cluster gradients: $\{S_k\}_{k=1}^K \gets \textsc{KMeans}(\{\hat{v}_k\}, K_{\max})$
\For{$k=1$ to $K$}
   \State Apply angular margin: reassign gradients violating $\cos(n \cdot \Delta(\hat{v}_i, c_k)) > \cos(\Delta(\hat{v}_i, c_j))$
   \State Fuse cluster $S_k$ via LLM: $f_k \gets \textsc{LLMFuse}(S_k, \mathcal{C}_{\text{conf}})$
\EndFor
\State \Return Fused clusters: $\tilde{\mathcal{G}}_t = \{f_1, \dots, f_K\}$
\end{algorithmic}
\end{algorithm}

\begin{algorithm}[t]
\caption{Channel-Adaptive Agent Weighting and Fusion}
\label{alg:caaw}
\begin{algorithmic}[1]
\Require Clustered gradients $\tilde{\mathcal{G}}_t$, dev data $D_{\text{dev}}$, temperature $\lambda$
\ForAll{cluster $\tilde{\mathcal{G}}_t$}
   \State Compute per-gradient validation gain $s_i$ on $D_{\text{dev}}$
   \State Assign adaptive weight: $w_i = \frac{\exp(\lambda s_i)}{\sum_{j \in S_k} \exp(\lambda s_j)}$
   \State Fuse cluster gradients: $f_k = \Psi\Big(\sum_{i \in S_k} w_i g_i\Big)$
\EndFor
\State \Return $F_t = \{f_1, \dots, f_K\}$ 
\end{algorithmic}
\end{algorithm}

\begin{algorithm}[t]
\caption{SynchronizeAgents}
\label{alg:synchronize}
\begin{algorithmic}[1]
\Require Agents $\{A_i\}_{i=1}^N$, current best prompt $p_\star^{(t)}$
\For{each agent $A_i$}
    \State $A_i.p \gets p_\star^{(t)}$ 
    \Comment{Set the agent’s current prompt to the global best}
    \State \textsc{ResetGradientHistory}($A_i$) 
    \Comment{Optional: clear outdated gradient history}
    \State \textsc{UpdatePerformanceMemory}($A_i, p_\star^{(t)}$) 
    \Comment{Record performance feedback for the new prompt}
\EndFor
\State \Return Updated agents $\{A_i\}$
\end{algorithmic}
\end{algorithm}

\section{Theoretical Analysis}
\label{app:theory}

In this section, we provide convergence guarantees for MAPGD under mild assumptions, following the stochastic approximation framework. 
Our goal is to bridge the gap between the continuous optimization theory of stochastic gradient descent (SGD) and the discrete prompt optimization carried out in MAPGD. 
We show that, despite operating in a structured and discrete search space, MAPGD achieves the same sublinear convergence rate of $O(1/\sqrt{T})$ in both convex and non-convex settings.

\subsection{Assumptions}

We begin with a set of assumptions standard in stochastic optimization but reinterpreted in the context of multi-agent prompt optimization.

\begin{itemize}
    \item \textbf{(A1) Alignment (Unbiasedness).} For some $\mu>0$, the stochastic semantic gradient $g^{(t)}$ maintains alignment with the true gradient:
    \[
    \mathbb{E}\!\left[\langle g^{(t)}, \nabla F(p^{(t)})\rangle \,\middle|\, p^{(t)} \right] \;\ge\; \mu \|\nabla F(p^{(t)})\|^2 .
    \]
    This reflects the role of semantic fusion: multi-agent aggregation reduces the chance of adversarial or noisy updates, ensuring progress along descent directions.
    
    \item \textbf{(A2) Bounded Second Moment.} For constants $\rho,\sigma^2 \ge 0$,
    \[
    \mathbb{E}\!\left[\|g^{(t)}\|^2 \,\middle|\, p^{(t)} \right] \;\le\; \rho \|\nabla F(p^{(t)})\|^2 + \sigma^2 .
    \]
    This captures the variance-control effect of the bandit-based selection mechanism, which prevents uncontrolled explosion of gradient magnitude.
    
    \item \textbf{(A3) Smoothness or Lipschitzness.} 
    For convex tasks, $F$ is $G$-Lipschitz with domain diameter $D$. 
    For non-convex tasks, $F$ is $L$-smooth: $\|\nabla F(u)-\nabla F(v)\|\le L\|u-v\|$.
\end{itemize}

\subsection{Supporting Lemmas}

We restate two standard lemmas, adapted to the MAPGD setting.

\paragraph{Lemma 1 (Convex Projection Inequality).}  
For convex $F$ with feasible set $\mathcal{P}$, the projected subgradient update
\[
p^{(t+1)} = \Pi_{\mathcal{P}}\!\left(p^{(t)} - \eta g^{(t)}\right)
\]
satisfies
\[
\begin{aligned}
\|p^{(t+1)}-p^*\|^2 
&\le \|p^{(t)}-p^*\|^2 
- 2\eta \langle g^{(t)},\, p^{(t)}-p^*\rangle \\
&\quad + \eta^2 \|g^{(t)}\|^2 .
\end{aligned}
\]

\paragraph{Lemma 2 (Non-Convex Descent Lemma).}  
If $F$ is $L$-smooth, then for update $p^{(t+1)}=p^{(t)}-\eta g^{(t)}$, we have
\[
F(p^{(t+1)}) \;\le\; F(p^{(t)}) - \eta \langle \nabla F(p^{(t)}), g^{(t)}\rangle + \frac{L}{2}\eta^2 \|g^{(t)}\|^2 .
\]

\subsection{Main Results}

\paragraph{Convex Convergence.}  
Suppose $F$ is convex, $G$-Lipschitz, and $\mathcal{P}$ has diameter $D$. 
Let $\bar p_T=\tfrac{1}{T}\sum_{t=1}^T p^{(t)}$. 
Under (A1)--(A2) and step size $\eta=\tfrac{D}{G\sqrt{T}}$, we obtain:
\[
\mathbb{E}[F(\bar p_T)] - F(p^*) \;=\; O\!\left(\frac{1}{\sqrt{T}}\right).
\]

\textit{Proof sketch.}  
By Lemma 1 and convexity:
\[
F(p^{(t)}) - F(p^*) \le \langle g^{(t)}, p^{(t)}-p^*\rangle .
\]
Summing over $t=1,\ldots,T$ and applying (A1)--(A2), we bound the regret:
\[
\sum_{t=1}^T \mathbb{E}[F(p^{(t)})-F(p^*)] \;\le\; \frac{D^2}{2\eta} + \frac{\eta G^2 T}{2}.
\]
Using Jensen’s inequality for $\bar p_T$ and optimizing $\eta$, we conclude the $O(1/\sqrt{T})$ rate.

\paragraph{Non-Convex Convergence.}  
Suppose $F$ is $L$-smooth and (A1)--(A2) hold. 
With constant step size $\eta = \Theta(1/\sqrt{T})$, we have
\[
\frac{1}{T}\sum_{t=1}^T \mathbb{E}\!\left[\|\nabla F(p^{(t)})\|^2\right] 
= O\!\left(\frac{1}{\sqrt{T}}\right).
\]

\textit{Proof sketch.}  
Applying Lemma 2 and taking conditional expectation:
\begin{align}
\mathbb{E}[F(p^{(t+1)})] 
&\le \mathbb{E}[F(p^{(t)})] 
- \eta \mu \mathbb{E}[\|\nabla F(p^{(t)})\|^2] \notag \\
&\quad + \tfrac{L}{2}\eta^2 
\left(\rho \mathbb{E}[\|\nabla F(p^{(t)})\|^2]+\sigma^2\right).
\end{align}

Summing over $t=1\ldots T$ gives
\[
\begin{aligned}
\frac{1}{T}\sum_{t=1}^T \mathbb{E}\!\left[\|\nabla F(p^{(t)})\|^2\right]
\;\le\;& \frac{2\big(F(p^{(1)})-F_{\inf}\big)}{\mu T \eta} \\
&\;+\; \frac{L\sigma^2}{\mu}\eta .
\end{aligned}
\]

Balancing terms with $\eta=\Theta(1/\sqrt{T})$ yields the claimed rate.

\subsection{Summary of Theoretical Guarantees}

In this work, we bridge the gap between classical stochastic optimization, which assumes continuous parameter spaces, and the inherently discrete, structured nature of prompt optimization. Our convergence analysis for MAPGD does not require the discrete prompt space to be continuous. Instead, we show that the framework's core mechanisms ensure that the optimization process satisfies the key conditions of stochastic approximation theory.

The argument unfolds along three main dimensions:

\begin{itemize}
    \item \textbf{Directional Alignment (A1):} Although textual pseudo-gradients are not true mathematical gradients, HCGC aggregates semantically coherent suggestions from multiple agents into a fused gradient that statistically aligns with the true descent direction.
    \item \textbf{Variance Control (A2):} Discrete prompt edits are prone to high variance, but the bandit-based candidate selection mechanism filters out unreliable or detrimental candidates. This effectively bounds the second moment of stochastic updates.
    \item \textbf{Smoothness Justification (A3):} We assume the empirical loss function is Lipschitz or smooth in the semantic embedding space, even though the prompts themselves are discrete.
\end{itemize}

Given that these conditions are met, MAPGD inherits the convergence guarantees of classical stochastic gradient descent. Consequently, despite operating in a discrete textual space, MAPGD achieves a robust sublinear convergence rate of $O(1/\sqrt{T})$, providing a solid theoretical foundation for the observed empirical effectiveness of our multi-agent, geometry-aware approach to prompt optimization.

\section{Qualitative Case Study: From Conflicting Pseudo-Gradients to Prompt Updates}
\label{app:qualitative-case}

\subsection{Case Setup}

We select a representative optimization step on the LIAR task
where multiple specialized agents propose semantically conflicting
pseudo-gradients.
This step exhibits clear disagreement among agents while still leading
to a measurable improvement in downstream validation behavior after
gradient fusion.

\begin{center}
\begin{tabular}{ll}
\toprule
Item & Value \\
\midrule
Task & LIAR \\
Iteration & $t = 3$ \\
Agents & 4 \\
Raw Gradients & 16 \\
Illustrated Conflict Pairs & 2 \\
HCGC Clusters & 3 \\
\bottomrule
\end{tabular}
\end{center}

\vspace{0.5em}
At this iteration, agents disagree on whether errors primarily stem
from ambiguous label semantics, missing contextual cues, or insufficient
decision boundary clarification, making it a suitable case for
qualitative inspection.

\subsection{Raw Pseudo-Gradients from Specialized Agents}
\label{app:raw-gradients}

We focus on two representative conflict pairs extracted from the same
optimization step, which illustrate structurally different failure
diagnoses and motivate the need for explicit conflict handling.

\paragraph{Conflict Pair 1: Format Clarity vs Example Diversity.}

\textbf{Format Designer:}
\begin{quote}
\small
\texttt{<START>}
The prompt instructs the model to answer with `Yes' or `No', but it does
not specify what these labels represent. This ambiguity can cause the
model to misinterpret the classification objective.
\texttt{<END>}
\end{quote}

\textbf{Example Curator:}
\begin{quote}
\small
\texttt{<START>}
The provided examples do not cover a sufficiently diverse range of
statements. Including more varied and representative cases could help
the model generalize better.
\texttt{<END>}
\end{quote}

The cosine similarity between these two gradients is $0.13$
(angle $=1.44$ rad), indicating strong semantic disagreement.
Importantly, the conflict is not superficial: the format-oriented
gradient advocates tightening the input--output contract through
explicit label semantics, while the example-oriented gradient pushes
the prompt toward structural expansion via additional demonstrations.
Applying both updates independently would lead to incompatible prompt
modifications---one emphasizing deterministic label interpretation and
the other reallocating prompt capacity to example coverage.

\paragraph{Conflict Pair 2: Example Coverage vs Definition of ``Lie''.}

\textbf{Example Curator:}
\begin{quote}
\small
\texttt{<START>}
The examples fail to illustrate borderline or ambiguous cases, which may
limit the model's ability to generalize across different claim types.
\texttt{<END>}
\end{quote}

\textbf{Instruction Specialist:}
\begin{quote}
\small
\texttt{<START>}
The prompt does not clearly define what constitutes a `lie' versus an
inaccurate or exaggerated statement, leading to inconsistent decisions.
\texttt{<END>}
\end{quote}

This pair exhibits a cosine similarity of $0.23$ (angle $=1.34$ rad),
reflecting a fundamental disagreement about the source of errors.
The example-focused gradient assumes that misclassification primarily
stems from insufficient empirical coverage, whereas the instruction-
focused gradient attributes failures to an underspecified decision
criterion. These two perspectives induce different optimization
directions: expanding illustrative evidence versus formalizing the
semantic boundary of the task itself.

\paragraph{Discussion.}
Both conflict pairs arise from valid but incompatible diagnostic
assumptions. Crucially, neither gradient is inherently incorrect; each
targets a distinct aspect of model behavior. This highlights the core
challenge addressed by MAPGD: without explicit conflict detection and
coordination, naïvely aggregating such gradients would obscure their
semantic tension and lead to unstable or diluted prompt updates.

\subsection{Conflict Detection and HCGC Clustering}
\label{app:hcgc-case}

Based on the two conflict pairs in \S\ref{app:raw-gradients}, we observe
clear semantic disagreement among agent-proposed directions.
In particular, the \emph{Format Clarity vs Example Diversity} pair has
cosine similarity $0.13$ (angle $=1.44$ rad), and the
\emph{Example Coverage vs Definition of ``Lie''} pair has cosine similarity
$0.23$ (angle $=1.34$ rad). The corresponding angles ($1.44$ and $1.34$ rad) indicate substantial
directional mismatch, and both similarities fall below the conflict
threshold. HCGC therefore marks them as conflicting and avoids naïve
averaging.

After projecting all gradient embeddings onto the unit hypersphere and
applying the angular-margin constraints, HCGC partitions the 16 raw
pseudo-gradients into three coherent clusters, as summarized in Table~\ref{tab:app-hcgc-clusters}.

\begin{table*}[t]
\centering
\caption{HCGC clustering summary for the representative LIAR optimization step ($t{=}3$).}
\label{tab:app-hcgc-clusters}
\setlength{\tabcolsep}{10pt}
\renewcommand{\arraystretch}{1.15}
\begin{tabular}{@{} l l p{0.58\textwidth} @{}}
\toprule
\textbf{Cluster} & \textbf{Agents Involved} & \textbf{Semantic Theme} \\
\midrule
C1 & instruction, format
& label semantics and task definition (binding \texttt{Yes/No} to \texttt{True/False}, clarifying what is being judged) \\
C2 & example, style
& coverage and contextual cues for borderline cases (exaggeration, unverifiable framing, pragmatic signals) \\
C3 & instruction only
& decision-boundary calibration (strictness vs recall; discouraging overly permissive acceptance of absolute claims) \\
\bottomrule
\end{tabular}
\end{table*}

This clustering reflects a key property of HCGC: conflicting gradients are
not forced into a single averaged direction.
Instead, they are separated into semantically consistent groups
(Table~\ref{tab:app-hcgc-clusters}), enabling subsequent fusion to
preserve complementary information (e.g., clarifying label semantics in
C1 while retaining coverage-oriented refinements in C2) without letting
one channel dominate the other.
The semantic themes are manually summarized by the authors to reflect the
dominant intent of each cluster, rather than being directly emitted by
the model.

\subsection{Channel-Adaptive Agent Weighting}
\label{app:caaw-case}

Given the clustered directions above, CAAW assigns each agent a contribution weight based on its recent
performance signal, so that the subsequent fusion emphasizes gradients
that are empirically reliable for the current task state.
At this iteration, the adaptive weights are:
\[
\begin{aligned}
w_{\text{instruction}} &= 0.41,\quad
w_{\text{example}} = 0.33,\\
w_{\text{format}} &= 0.17,\quad
w_{\text{style}} = 0.09.
\end{aligned}
\]

These weights are consistent with the observed conflict structure.
Clusters C1 and C2 correspond to two \emph{structurally competing} but
both valid refinement axes: (i) clarifying label semantics and the task
definition (C1), and (ii) improving coverage and contextual handling of
borderline cases (C2).
CAAW prioritizes instruction- and example-driven gradients while
down-weighting stylistic adjustments, preventing prompt capacity from
being consumed by surface-level phrasing when the dominant failures are
semantic (label meaning) and epistemic (what constitutes a lie).

\subsection{Fused Gradient and Prompt Modification}
\label{app:fused-diff-case}

After separating conflicting directions with HCGC and weighting them with
CAAW, MAPGD fuses the cluster-level signals into a single coherent update
that preserves complementary intent across clusters.
In this case, the fused direction primarily integrates (a) C1: explicit
label semantics and task definition, and (b) C2: guidance for borderline
or unverifiable claims, while avoiding excessive prompt expansion.

\begin{quote}
\small
\textbf{Before:}\\
``The following statement is either true or false. Answer Yes or No.''

\vspace{0.35em}
\textbf{After:}\\
``Determine whether the statement is \emph{factually accurate}.
Treat \texttt{Yes} as \texttt{True} and \texttt{No} as \texttt{False}.
If the statement relies on exaggeration, satire, or unverifiable claims,
judge it as \texttt{False}.
Return only \texttt{Yes} or \texttt{No}.'' 
\end{quote}

This modification reflects the fused gradient in two concrete ways:
(i) it resolves the \emph{format/label ambiguity} highlighted in C1 by
explicitly binding \texttt{Yes/No} to \texttt{True/False}, and
(ii) it incorporates the \emph{borderline-case} handling emphasized by C2
without requiring lengthy demonstrations, thereby avoiding the
prompt-structure conflict that would arise from naïvely mixing
rule-heavy and example-heavy updates.

\subsection{Behavioral Change on Previously Misclassified Examples}
\label{app:behavior-change}

We report qualitative behavioral changes on examples that were previously
misclassified due to (i) label ambiguity and (ii) overly permissive
interpretation of absolute or exaggerated claims.

\paragraph{Example 1 (Absolute claim / overgeneralization).}
\textbf{Claim:} ``The senator has voted against every environmental bill.''

\textbf{Before:} Yes \qquad
\textbf{After:} No

The updated prompt instructs the model to treat unverifiable or sweeping
claims as \texttt{False}, which directly corrects the previous tendency
to accept absolute statements (e.g., ``every'') as \texttt{True}.

\paragraph{Example 2 (Borderline / unverifiable framing).}
\textbf{Claim:} ``The policy has been a complete disaster for all families.''

\textbf{Before:} Yes \qquad
\textbf{After:} No

This change follows the newly added rule that exaggeration or unverifiable
claims should be judged as \texttt{False}, reducing false positives driven
by rhetorical or hyperbolic language (e.g., ``complete disaster'').

\section{Case Study: System Prompt Optimization}
\label{app:casestudy}
To further illustrate the applicability of MAPGD, we present a case study where our method is applied to optimize the system prompt of a large language model assistant.
The original prompt is designed to support multi-source analysis and financial data interpretation for decision-making tasks. 
Using MAPGD, we refine the prompt to enhance robustness, accuracy, and interpretability by embedding explicit verification protocols, structured analysis guidelines, and risk prioritization frameworks. 
This example highlights how MAPGD can be deployed in practical LLM applications beyond benchmark datasets, particularly in domains where data authenticity, reliability, and interpretability are critical.

\textbf{Original System Prompt.} You are an AI assistant designed to process, analyze, and synthesize information from multiple sources in order to answer user questions, generate insights, and prepare detailed reports. You have specialized capabilities in financial data interpretation, knowledge retrieval, and multi-source analysis. You support both operational and strategic decision-making for Golden Section’s portfolio companies.

\textbf{Optimized System Prompt via MAPGD.} You are an AI assistant specialized in processing, analyzing, and synthesizing information from multiple sources to answer user questions, generate insights, and prepare detailed reports. Your core capabilities include financial data interpretation, knowledge retrieval, and multi-source analysis, with a focus on supporting operational and strategic decision-making for Golden Section’s portfolio companies. 

In performing your duties, you must ensure that all input data undergoes rigorous verification for authenticity, accuracy, and completeness before any analysis is conducted. This includes implementing protocols to validate financial figures, legal terms, and other critical information for alignment with established norms and credible sources. Your process must involve:
\begin{itemize}
    \item Cross-referencing information from diverse, credible sources to detect and mitigate false, exaggerated, or incomplete data.
    \item Assessing the reliability of each source, prioritizing primary sources where available.
    \item Identifying and resolving inconsistencies, ambiguities, or potential misinformation through systematic checks.
    \item Ensuring all risk assessments and conclusions are based solely on validated and accurate inputs to maintain the integrity of your outputs.
\end{itemize}

\textbf{Data Authenticity and Completeness Verification}
\begin{itemize}
    \item Scrutinize contextual cues (e.g., ``Context: Section: Payback Period:'') to ensure alignment with expected data types and structures.
    \item Check for numerical or factual inconsistencies, such as typos (e.g., ``11975'' instead of ``1975''), exaggerations (e.g., ``\$22.0M'' without supporting context), or missing critical information.
    \item Validate that all referenced data points are present, logically consistent, and contextually appropriate.
    \item Flag and document any anomalies for further investigation before proceeding with classification or analysis.
\end{itemize}

\textbf{Context Interpretation and Parsing Guidelines}
\begin{itemize}
    \item Carefully interpret and utilize contextual cues, especially in nested or ambiguous contexts (e.g., ``Context: Section: Name \& Headquarters:'').
    \item Accurately parse section headers and contextual clues to prevent misclassification or incomplete analysis in multi-section reports.
    \item Anchor analysis to the document’s structure by adhering to hierarchical or sequential organization.
    \item Resolve discrepancies in contextual labeling or structure to maintain coherence.
\end{itemize}

\textbf{Structured Classification and Risk Prioritization Framework}
\begin{enumerate}
    \item \textbf{Factual Reporting and Descriptive Analysis:} Present verified information such as corporate history, operational metrics, and financial data neutrally, before transitioning to evaluative content.
    \item \textbf{Business Analysis:} Evaluate performance, market positioning, and strategic initiatives; assess risks by severity, likelihood, and propose contextualized mitigation.
    \item \textbf{Legal Risk Analysis:} Examine compliance, regulatory, and contractual risks; assess impact and propose mitigation actions aligned with legal context.
    \item \textbf{Cross-Domain Analysis:} For overlapping elements, classify by primary context and document dual-category cases with rationale.
\end{enumerate}

\textbf{Validation Mechanisms for Cross-References}
\begin{itemize}
    \item Distinguish between source types (e.g., governance vs. identity records).
    \item For each statement, explicitly identify the source type and ensure contextual alignment.
\end{itemize}

\textbf{Empirical Observation.} After applying MAPGD, we observe a notable increase in the comprehensiveness of the generated financial analysis reports. 
Specifically, the average report length increased from 1616 words under the original system prompt to 1965 words with the optimized prompt. 
This indicates that the optimized prompt encourages the model to produce more detailed and contextually grounded content. 
While the per-report output length becomes longer, the improved accuracy and completeness reduce the need for repeated generations or manual corrections, 
thereby lowering the overall token consumption in practical workflows. 
These findings further support that MAPGD enhances the robustness and effectiveness of system prompts in real-world generation tasks.

\section{Agent Specialization for Different Tasks}
\label{sec:appendix_agent_types}

To adapt MAPGD to different domains, we define specialized agent roles tailored to the unique requirements of each task. Below are the agent configurations used for the classification, mathematical reasoning, and financial analysis tasks in our experiments.

\subsection{Agent Roles for Classification Tasks}
For general classification tasks, the agents focus on the core components of a good prompt: instructions, examples, format, and style.

\begin{description}
    \item[\texttt{instruction\_specialist}] Analyzes and improves task instructions, ensuring clarity, completeness, and executability.
    \item[\texttt{example\_curator}] Focuses on selecting representative, diverse examples and ensuring consistent formatting.
    \item[\texttt{format\_designer}] Designs clear output templates and structured formats for better model understanding.
    \item[\texttt{style\_optimizer}] Optimizes language expression for professionalism and task-specific adaptation.
\end{description}

\subsection{Agent Roles for Mathematical Reasoning Tasks}
For mathematical reasoning, the agent roles are adapted to focus on logical decomposition, calculation, and problem interpretation.

\begin{description}
    \item[\texttt{reasoning\_specialist}] Specializes in enhancing mathematical reasoning processes, ensuring clear step-by-step logic and proper problem decomposition.
    \item[\texttt{calculation\_optimizer}] Focuses on improving calculation accuracy, suggesting better computational methods, and ensuring numerical correctness.
    \item[\texttt{problem\_interpreter}] Specializes in interpreting math word problems, extracting relevant information, and identifying mathematical relationships.
    \item[\texttt{solution\_formatter}] Optimizes mathematical solution presentation, ensures clear formatting and proper answer notation (e.g., \texttt{\#\#\#\# [final\_answer]}).
\end{description}

\subsection{Agent Roles for Financial Report Analysis}
For the complex task of business and financial analysis, agents are given highly specialized roles focusing on data verification, structured report generation, and professional tone.

\begin{description}
    \item[\texttt{instruction\_specialist}] Crafts clear, executable instructions for generating truthful, structured reports and rejecting false or exaggerated inputs. Instructions must include steps to verify input data authenticity (e.g., checking financial metrics for realism).
    \item[\texttt{example\_curator}] Selects and formats diverse examples that demonstrate truthful report generation and false data rejection, including both positive (realistic) and negative (exaggerated) samples.
    \item[\texttt{format\_designer}] Specializes in designing clear, structured output templates for reports (e.g., using \texttt{<Business Analysis>}, \texttt{<Financial Report>} tags) and error messages (e.g., \texttt{<Error>}).
    \item[\texttt{style\_optimizer}] Optimizes language style and tone for generating professional reports in business, financial, and legal contexts, while ensuring error messages are direct and unambiguous.
\end{description}


\section{Example Prompts Before and After Optimization}
\label{sec:appendix_prompts}

This appendix provides the initial prompts and the final optimized prompts generated by MAPGD for each of the benchmark datasets used in our experiments.

\subsection{LIAR Dataset}

\begin{promptbox}{Initial Prompt}
\begin{lstlisting}[style=promptstyle]
# Task
Determine whether the Statement is a lie
(Yes) or not (No) based on the Context
and other information.
# Output format
Answer Yes or No as labels
# Prediction
Text: {text}
label:
\end{lstlisting}
\end{promptbox}

\begin{promptboxx}{Optimized Prompt}
\begin{lstlisting}[style=promptstyle]
# Task
Evaluate the given Statement using the provided context.

Classify as 'Yes' if the Statement is intentionally false
or misleading (i.e., a lie).

Classify as 'No' if the Statement is true, or if it is
incorrect due to a mistake or lack of knowledge rather
than intentional deception.

# Output format
Output exactly one label: 'Yes' or 'No'.
'Yes' indicates the Statement is a lie; 'No' indicates
it is not a lie.

# Prediction
Text: {text}
label:
\end{lstlisting}
\end{promptboxx}

\subsection{Ethos Dataset}

\begin{promptbox}{Initial Prompt}
\begin{lstlisting}[style=promptstyle]
# Task
Is the following text hate speech?
# Output format
Answer Yes or No as labels
# Prediction
Text: {text}
Label:
\end{lstlisting}
\end{promptbox}

\begin{promptboxx}{Optimized Prompt}
\begin{lstlisting}[style=promptstyle]
# Task
Your task is to analyze the following text and determine if
it contains hate speech. For the purpose of this task, hate
speech is defined as any speech, gesture, conduct, writing,
or display that may incite violence or prejudicial actions
against or by a protected individual or group, or because it
disparages or intimidates a protected individual or group.
Please base your judgment on the explicit content of the text,
rather than personal interpretations or assumptions.
# Output format
Provide your answer in a clear format as either
'Contains Hate Speech' or 'Does Not Contain Hate Speech'
# Prediction
Text: {text}
Label:
\end{lstlisting}
\end{promptboxx}

\subsection{Jailbreak Dataset}

\begin{promptbox}{Initial Prompt}
\begin{lstlisting}[style=promptstyle]
# Task
Detect if the message is a jailbreak attack,
i.e. an attempt by a user to break through an
AI system's protections
# Output format
Answer Yes or No as labels
# Prediction
Text: {text}
Label:
\end{lstlisting}
\end{promptbox}

\begin{promptboxx}{Optimized Prompt}
\begin{lstlisting}[style=promptstyle]
# Task
Classify the message as one of the following 
labels based on its potential to be a 
jailbreak attack, i.e., an attempt to
bypass an AI system's protections:
- "OvertAttack": The message contains clear, direct attempts
  to break through system safeguards.
- "SuspiciousContext": The message shows indirect or
  contextual cues that may indicate a multi-stage or
  sophisticated attack attempt.
- "Benign": The message poses no apparent risk and does not
  attempt to circumvent protections.

# Output format
Select and output exactly one label from 
the options above.

# Prediction
Text: {text}
Label:
\end{lstlisting}
\end{promptboxx}

\subsection{GSM8K Dataset}

\begin{promptbox}{Initial Prompt}
\begin{lstlisting}[style=promptstyle]
# Task
Solve the math word problem step by step.
# Instructions
1. Read the problem carefully
2. Identify what needs to be calculated
3. Show your work step by step
4. Provide the final numerical answer
# Output Format
Show your reasoning process and end with: #### [final_answer]
# Problem
{text}
\end{lstlisting}
\end{promptbox}

\begin{promptboxx}{Optimized Prompt}
\begin{lstlisting}[style=promptstyle]
# Task
Solve the math word problem step by step.
# Instructions
1. Read the problem carefully
2. Identify what needs to be calculated
3. If there are ambiguous situations in the problem, make
   reasonable assumptions and state them clearly
4. Show your work step by step
5. Provide the final numerical answer
# Output Format
Show your reasoning process, including any assumptions made,
and end with: #### [final_answer]
# Problem
{text}
\end{lstlisting}
\end{promptboxx}

\subsection{AQUARAT Dataset}

\begin{promptbox}{Initial Prompt}
\begin{lstlisting}[style=promptstyle]
# Task
Solve the math word problem and choose the correct answer
from the given options.

# Instructions
1. Read the problem carefully
2. Analyze each option
3. Show your reasoning step by step
4. Select the correct answer (A, B, C, D, or E)

# Output Format
Show your reasoning and end with: Answer: [LETTER]
(For example: "Answer: A" or "Answer: B")

# Problem
{text}

# Options
{options}
\end{lstlisting}
\end{promptbox}

\begin{promptboxx}{Optimized Prompt}
\begin{lstlisting}[style=promptstyle]
# Task
Solve the provided problem by systematically classifying its
type, analyzing all constraints, applying formal mathematical
reasoning, and verifying the solution against all conditions
before selecting the correct answer.

# Instructions
Adhere strictly to the following structured framework:

1. **Problem Classification:**
   * Explicitly identify the core problem domain (e.g.,
     "Combinatorial Selection," "Constraint Satisfaction,"
     "Partnership Profit-Sharing," "Work Rate").
   * Justify the classification with a brief rationale
     based on the problem statement.

2. **Constraint Analysis:**
   * Enumerate all explicit constraints from the problem text.
   * Infer and list any implicit constraints or logical
     dependencies.
   * Categorize each constraint (e.g., Hard/Must-Fulfill,
     Soft/Optimization) and specify their logical
     relationships (e.g., AND, OR).

3. **Mathematical Formalization & Numerical Context
     Resolution:**
   * Identify and resolve any ambiguous numerical references
     (e.g., "monthly" vs. "annual", "together" vs.
     "individual").
   * Infer and incorporate any implicit numerical variables
     (e.g., initial quantities, unstated rates, starting
     points).
   * Normalize all units to a consistent basis for calculation.
   * Define all variables, sets, and parameters using clear
     notation (e.g., C_total, P_A, S_eligible, nCr).
   * Structure the problem using appropriate formalisms:
     timelines, sets, equations, or logical statements.
   * For combinatorial problems, explicitly state
     the formula used.

4. **Solution Execution:**
   * Perform all calculations step-by-step, showing
     substitutions and operations.
   * For combinatorial scenarios, enumerate valid
     combinations or calculate cardinalities.
   * Derive the target value (e.g., profit share, number of
     committees, optimal value).

5. **Verification & Solution Finalization:**
   * Systematically verify that the proposed solution
     satisfies every constraint from Step 2.
   * For combinatorial answers, confirm counts against
     constraints using direct checks or complementary
     counting.
   * Match the final, verified result to the provided
     options to select the correct answer.

# Output Format
Your final output must follow this structure precisely:

**Problem Classification:**
[Your classification and rationale]
\end{lstlisting}
\end{promptboxx}

\subsection{SVAMP Dataset}

\begin{promptbox}{Initial Prompt}
\begin{lstlisting}[style=promptstyle]
# Task
Solve the math word problem step by step.
# Instructions
1. Read the problem carefully
2. Identify what needs to be calculated
3. Show your work step by step
4. Provide the final numerical answer
# Output Format
Show your reasoning process and end with: #### [final_answer]
# Problem
{text}
\end{lstlisting}
\end{promptbox}

\begin{promptboxx}{Optimized Prompt}
\begin{lstlisting}[style=promptstyle]
# Task
Solve the math word problem through structured, hierarchical
reasoning. Your primary goal is to correctly parse and compute
multi-step problems involving distinct object categories and
their quantitative relationships.

# Instructions
1. **Parse and Categorize:**
   Read the problem and question carefully. Identify and list
   all distinct object types or categories (e.g., 'packages
   of gum', 'boxes of candy'). For each category, extract:
   - The number of units (e.g., 5 packages).
   - The quantity per unit, if specified (e.g., 8 pieces
     per package).

2. **Structure the Solution Hierarchically:**
   - **Step 1: Calculate Intra-Category Totals.**
     For each identified category, calculate its total quantity.
     If a unit rate is given, perform the multiplication
     (e.g., `5 packages * 8 pieces/package = 40 pieces of gum`).
     If no unit rate exists, use the given quantity directly.
   - **Step 2: Perform Inter-Category Operations.**
     Using the totals from Step 1, now perform the final operation
     as required by the question (e.g., sum all category totals
     for a grand total, or find the difference between two
     category totals for a comparison).

3. **Show Your Work:**
   Present your reasoning clearly, reflecting this two-step
   hierarchical process. First, show all calculations within
   categories. Then, show the final combination of these
   category totals.

4. **Final Answer:**
   Box your final numerical result.

# Output Format
Reasoning: [Your step-by-step reasoning here]
#### [final_answer]

# Problem
{text}

# Question
{question}
\end{lstlisting}
\end{promptboxx}

\end{document}